%% file: main.tex
\documentclass[10pt,twocolumn,letterpaper]{article}

\usepackage[pagenumbers]{cvpr} 

\input{preamble}

\definecolor{cvprblue}{rgb}{0.21,0.49,0.74}
\usepackage[pagebackref,breaklinks,colorlinks,citecolor=cvprblue]{hyperref}

\usepackage[utf8]{inputenc} 
\usepackage[T1]{fontenc}    

\usepackage{graphicx}
\usepackage{amsmath}
\usepackage{amssymb}
\usepackage{bbm}

\usepackage{cite}
\usepackage{amsmath,amssymb,amsfonts}
\usepackage{algorithmic}
\usepackage{textcomp}
\usepackage{subcaption}
\usepackage{tabularx}
\usepackage{booktabs}       
\usepackage{amsmath}
\usepackage{amssymb}
\usepackage{bm}
\usepackage{color}
\usepackage{xcolor,soul}
\usepackage{algorithm}
\usepackage{multirow}
\usepackage{xspace}
\usepackage{soul}

\usepackage{enumitem}

\usepackage{newtxtext}
\newdimen\abovecrulesep
\newdimen\belowcrulesep
\makeatletter
\patchcmd{\@@@cmidrule}{\aboverulesep}{\abovecrulesep}{}{}
\patchcmd{\@xcmidrule}{\belowrulesep}{\belowcrulesep}{}{}

\definecolor{demphcolor}{RGB}{144, 144, 144}
\definecolor{mygray}{gray}{0.4}
\definecolor{lightgray}{rgb}{0.9, 0.9, 0.9}
\definecolor{deepgreen}{RGB}{0,100,0}
\newcommand{\demph}[1]{\textcolor{demphcolor}{#1}}

\newlength\savewidth
\newcommand\shline{\noalign{\global\savewidth\arrayrulewidth\global\arrayrulewidth 1pt}\hline\noalign{\global\arrayrulewidth\savewidth}}

\makeatletter\renewcommand\paragraph{\@startsection{paragraph}{4}{\z@}{.5em\@plus1ex\@minus.2ex}{-.5em}{\normalfont\normalsize\bfseries}}
\makeatother

\newcommand{\modelname}{HACL\xspace}

\setstcolor{blue}

\usepackage[capitalize]{cleveref}
\crefname{section}{Sec.}{Secs.}
\Crefname{section}{Section}{Sections}
\Crefname{table}{Table}{Tables}
\crefname{table}{Tab.}{Tabs.}

\def\confName{CVPR}
\def\confYear{2024}

\title{Hallucination Augmented Contrastive Learning for Multimodal Large Language Model }

 \author{
Chaoya Jiang$^1$\hspace{6mm}  Haiyang Xu$^2$\thanks{Corresponding author}\hspace{6mm} Mengfan Dong$^1$\hspace{6mm}
Jiaxing Chen$^1$\hspace{6mm} Wei Ye$^1$\footnotemark[1]\hspace{6mm} \\ Ming Yan$^2$\hspace{6mm} Qinghao Ye$^2$\hspace{6mm} Ji Zhang$^2$\hspace{6mm} Fei Huang$^2$\hspace{6mm}  Shikun Zhang$^1$\\
$^1$National Engineering Research Center for Software Engineering, Peking University \\
$^2$Alibaba Group \\
{\tt\small \{jiangchaoya, wye\}@pku.edu.cn, shuofeng.xhy@alibaba-inc.com} \\
}
\begin{document}
\maketitle
\begin{abstract}
Multi-modal large language models (MLLMs) have been shown to efficiently integrate natural language with visual information to handle multi-modal tasks. However, MLLMs still face a fundamental limitation of hallucinations, where they tend to generate erroneous or fabricated information. In this paper, we address hallucinations in MLLMs from a novel perspective of representation learning. We first analyzed the representation distribution of textual and visual tokens in MLLM, revealing two important findings: 1) there is a significant gap between textual and visual representations, indicating unsatisfactory cross-modal representation alignment; 2) representations of texts that contain and do not contain hallucinations are entangled, making it challenging to distinguish them. These two observations inspire us with a simple yet effective method to mitigate hallucinations. Specifically, we introduce contrastive learning into MLLMs and use text with hallucination as hard negative examples, naturally bringing representations of non-hallucinative text and visual samples closer while pushing way representations of non-hallucinating and hallucinative text. We evaluate our method quantitatively and qualitatively, showing its effectiveness in reducing hallucination occurrences and improving performance across multiple benchmarks. On the MMhal-Bench benchmark, our method obtains a 34.66\% /29.5\% improvement over the baseline MiniGPT-4/LLaVA. Our code is available on \url{https://github.com/X-PLUG/mPLUG-HalOwl/tree/main/hacl}.

\end{abstract}

\section{Introduction}

\begin{figure}[h]
    \centering
     \vspace{-2ex}
    \includegraphics[width=1\linewidth]{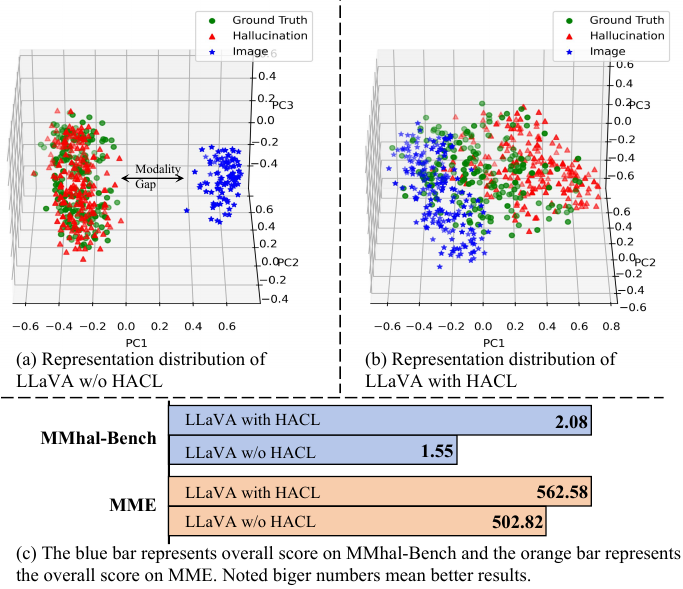}
    \caption{Subfigure (a) and subfigure (b) show the distributions of the last token's representations yielded by LLM for visual or textual token sequences. Blue icons represent images, green icons represent ground truth captions, and red ones represent hallucinative captions generated by GPT-4. HACL refers to our proposed method, Hallucination Augmented Contrastive Learning. In subfigure (a), textual and visual representations have cross-model semantic gaps, while non-hallucinating and hallucinative text representations are mixed. This phenomenon is alleviated by HACL, as shown in subfigure (b). Subfigure (c) shows the empirical results of the hallucination evaluation benchmark MMhal-Bench \cite{Sun2023LLavaRlhf} and the model performance evaluation metric MME \cite{fu2023mme}.}
    \vspace{-2ex}
    \label{fig:fig1}
\end{figure}

Large Language Models (LLMs) like GPT-3 \cite{Brown2020gpt3}, LLaMA \cite{Touvron2023LLaMA, Touvron2023Llama2}, and GPT-4 \cite{OpenAI2023gpt4} have received significant attention for their remarkable text understanding and generation capabilities. Recently, GPT-4V\footnote{https://openai.com/research/gpt-4v-system-card} \cite{2023GPT4VisionSC} has demonstrated impressive multi-modal abilities in tasks such as image caption and visual question answering, shedding light on the vision-language domain and attracting widespread research interests.
Consequently, a new category of models, known as Multi-modal Large Language Models (MLLMs) \cite{Zhu2023MiniGPT4, Liu2023Llava, ye2023mplugowl, mplugdocowl, ye2023ureader, Bai2023QwenVL, Dai2023InstructBLIP, Li2023BLIP2}, has emerged, aiming to enhance LLMs with the capacity to comprehend and handle visual information. 
To integrate natural language with other modalities, MLLMs incorporate a learnable interface between pre-trained visual encoders and LLMs. Such interface includes learnable query tokens \cite{Li2023BLIP2, Dai2023InstructBLIP, Zhu2023MiniGPT4, ye2023mplugowl} or a projection-based linear model \cite{Liu2023Llava, Liu2023LLava15} that extracts and integrates information from visual modalities. MLLMs learn this interface to generate answers for multimodal instructions, resulting in remarkable performance in many multimodal tasks.

However, a fundamental limitation of MLLMs is their tendency to produce erroneous or fabricated information that doesn't match the provided visual input, known as hallucination \cite{Li2023pope, liu2023mitigating, Wang2023VIGCVI, Sun2023LLavaRlhf}. In this paper, we aim to tackle the issue from the perspective of representation learning. We first check the distribution of textual and visual tokens within the representation space of LLMs (Vicuna \cite{zheng2023vicuna} in our experiments), in which visual representations are projected by the learned interface. As shown in Figure \ref{fig:fig1}, we have two primary findings:
\begin{itemize}
    \item \textit{A significant modality gap remains between the textual and visual tokens despite visual projection;}
    \item \textit{Representations of texts that contain and do not contain hallucinations are entangled, making it challenging to differentiate them.}
\end{itemize}
These preliminary observations indicate that the current learned interfaces are not effective enough to map visual representations into the textual representation space of LLMs. As a result, it is difficult for MLLMs to discriminate between texts containing minor errors at the level of objects or attributes and those manifesting typical hallucinative expressions. This issue potentially heightens the tendency for MLLMs to generate more hallucinations. Therefore, exploring more effective approaches to align visual representations with LLMs' textual representation space to address hallucinations is crucial.

Inspired by the findings above, we propose hallucination-augmented cross-modal contrastive learning (\modelname), which enhances the alignment between visual and textual representations to alleviate hallucinations. Texts with hallucination are used as hard negative examples for image anchors, naturally pulling closer representations of non-hallucinating text and visual samples while pushing way representations of non-hallucinating and hallucinative text. Specifically, we separately feed the visual and textual token sequences into LLMs to obtain global representations for each modality, which is used for contrastive learning. We generate hallucinative image captions with GPT-4 \cite{OpenAI2023gpt4}. These hallucinative texts contain partial object attribute errors or introduce additional non-existent information compared to the original image captions. As shown in Figure \ref{fig:fig1} (b), introducing \modelname into LLaVA \cite{Liu2023Llava} forces the visual representation closer to the text representation and makes the correct and hallucinated text representations more distinguishable. This effective alignment helps to prevent the generation of hallucinations. Our experiments also show that equipping MLLMs with \modelname not only reduces the occurrence of hallucinations but also yields improvements across multiple benchmark evaluations. As shown in Subfigure \ref{fig:fig1} (c), when equipped with \modelname, LLaVA achieves a 29\% increase in overall score on the MMhal-Bench benchmark \cite{Sun2023LLavaRlhf}, as well as an 11\% improvement on the MME \cite{fu2023mme} benchmark. In conclusion, this paper makes the following contributions:
\begin{itemize}
\item We underline a significant cross-modal semantic gap between visual and textual representations and an unexpected representation tangling among text containing and not containing hallucinations in MLLMs.
These findings expose the inadequacies of current methodologies in efficiently bridging the gap between visual and textual representations.

\item Based on these insights, we propose a simple yet effective method named Hallucination Augmented Cross-Modal Contrastive Learning (\modelname). Introducing contrastive learning into MLLMs and using hallucinative text as hard negative samples yields a better cross-modal and more hallucinations-distinguishable representation space.

\item Our experiments show that equipping MLLMs with \modelname not only minigates hallucinations but also effectively improve the performance across multiple benchmark evaluations.

\end{itemize}

\section{Related Work}
\label{sec:related_work}
 
\paragraph{Multi-Modal Large Language Foundation Models.}
The successful application of Large Language Models (LLMs) has paved the way for developing several approaches aiming to augment the perceptual capacities of LLMs with additional modalities, all within a unified model. There are three primary methods for constructing multi-modal large language foundational models, each showing promise for robust zero-shot generalization capabilities in the vision-language domain. For instance, Flamingo \cite{alayrac2022flamingo} is a forerunner in this area, using a frozen vision encoder and a large language model equipped with gated cross-attention for cross-modality alignment. In contrast, PaLM-E \cite{Driess2023PaLME} integrates extracted visual features directly through linear layers into the pre-trained PaLM \cite{Chowdhery2022PaLM} model, which boasts 520 billion parameters, thereby leading to robust performance across numerous real-world applications. This approach has been broadly adopted by models such as LLaVA \cite{Liu2023Llava}, Shikra \cite{Chen2023Shikra}, etc. One significant limitation of this method, however, is the creation of lengthy visual sequences. To address this, BLIP-2 \cite{Li2023BLIP2}, drawing inspiration from DETR \cite{carion2020detr}, developed a Q-former to reduce the sequence length of visual features efficiently. This design has been mirrored by Kosmos-1 \cite{Huang2023Kosmos1}, mPLUG-Owl \cite{ye2023mplugowl}, and MiniGPT-4 \cite{Zhu2023MiniGPT4}. 

 \begin{figure*}[t]
    \centering
    \includegraphics[width=1\textwidth]{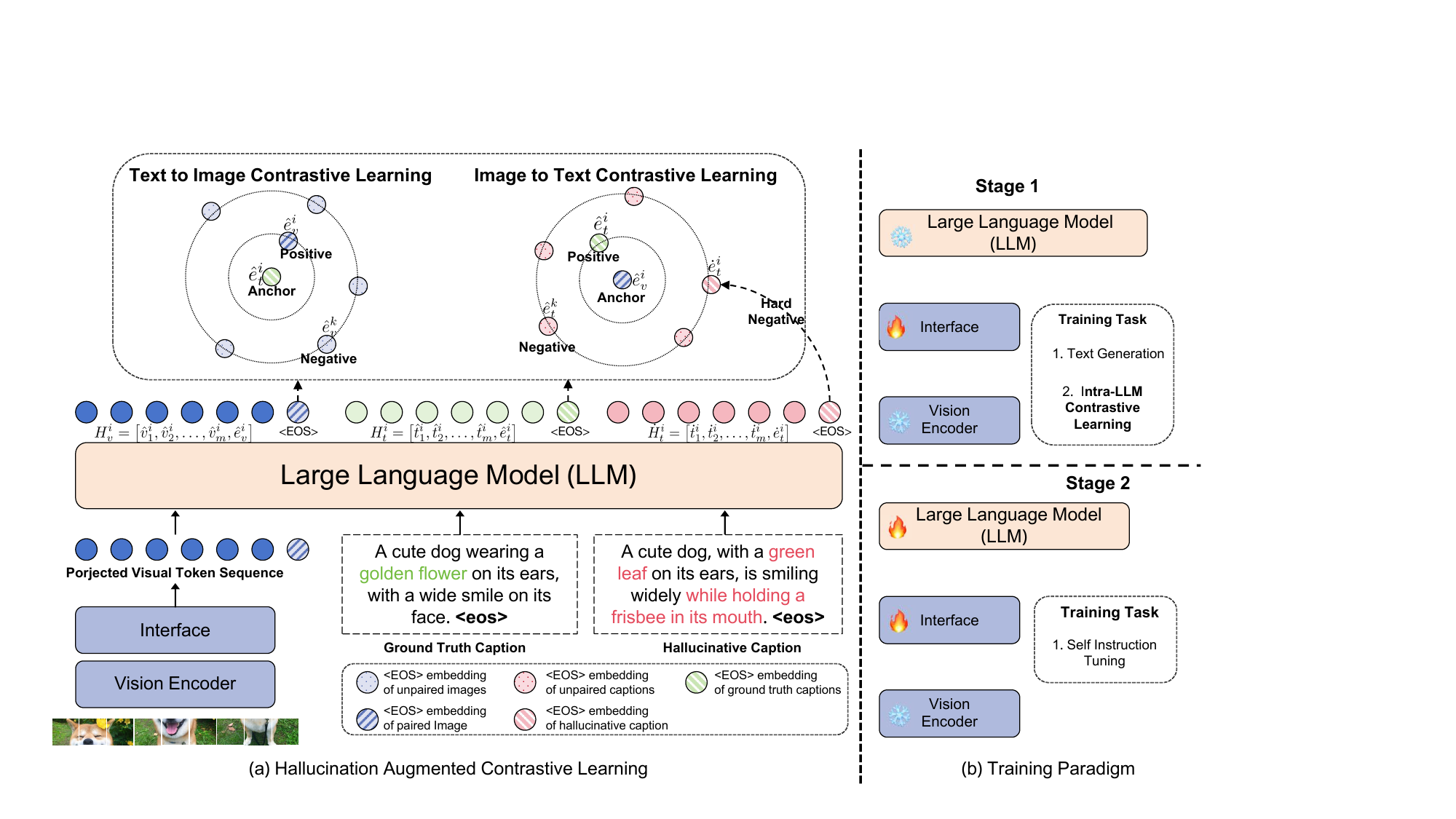}
\vspace{-3ex}
    \caption{Subfigure (a) illustrates the proposed \modelname. In this framework, we employ GPT-4 \cite{OpenAI2023gpt4} to generate the hallucinative captions as the the hard negative samples in the image-to-text contrastive learning. Subfigure (b) shows the training paradigm of \modelname. }
\label{fig:arch}
\end{figure*}

\paragraph{Minigating Hallucination for MLLMs.}
To address the issue of hallucination in MLLMs, researchers have developed various methods, which can be broadly categorized into two lines. The first line \cite{liu2023lrv, Wang2023VIGCVI} involve limiting the length of instruction data, which typically leads to a reduction in hallucination. For instance, LRV-Instruction\cite{liu2023lrv} takes an intuitive approach by constraining the text length of instructions and constructing counterfactual instructions. However, this may result in less detailed descriptions from the fine-tuned model. The second line utilizes additional artificial data or tools to modify hallucinations in the model's output. For example, LLaVA-RLHF \cite{Sun2023LLavaRlhf} employs manually annotated data as reward signals to guide the model in generating less hallucinative responses. Although effective, this approach requires extra manual annotation data. In this paper, we propose a method from the perspective of representation learning. We introduce hallucinative captions as hard negative samples in contrastive learning, aiming to narrow the gap between visual representations and correct textual representations, while pushing away from hallucinative textual representations. This approach effectively addresses the issue of hallucination and also enhances the model's visual understanding capability.

\section{Method}

The learnable interface of MLLMs plays a vital role in bridging diverse modalities and mapping visual representations to the representation space of LLMs. Our goal is to refine this interface to facilitate better matching of visual representations with the ground truth text in the representation space, while also increasing the distance between them and hallucinative text. To accomplish this, we propose a new approach called Hallucination Augmented Cross-modal Contrastive Learning (\modelname). This approach is inspired by contrastive learning, which is a well-established technique in the fields of representation learning \cite{oord2018representation} and self-supervised learning \cite{radford2021clip, he2020momentum, chen2020moco, Jiang2023VisionLP}. In the following subsection, we first introduce how to incorporate  cross-modal contrastive learning during training. Next, we describe how to boost contrastive learning through additional generated hallucinative captions. Finally, we introduce the hallucination augmented contrastive learning training paradigm.

\subsection{Cross-modal Contrastive Learning}
\label{subsec:ILCL}
As shown in  Figure \ref{fig:arch} (a),   our approach can be applied to any MLLMs that maps or abstracts visual information to the textual representation space through an learnable interface.  Formally, we assume that the MLLM consists of a vision encoder denoted as $\mathbf{V}_\theta$, a learnable interface denoted as $\mathbf{F}_\alpha$, and a decoder-only based Large Language Model denoted $\mathbf{L}_\beta$ where $\theta, \alpha, \beta$ represent the parameters of each module. Additionally, we also have an unsupervised pretraining dataset, containing N image-text pairs, denoted as $D=\{ I_i, T_i \}, i\in \left[1,2,\dots, N\right]$.

 Assuming an image $I_i$ is processed by the vision encoder $\mathbf{V}_\theta$ and the learnable interface $\mathbf{F}_\alpha$, it is transformed into a visual token sequence of length $m$. Since most LLMs are decoder-only models, in order to obtain the representations that can capture global semantic information. We pass a $<EOS>$ token through an embedding layer $\mathbf{L}_\beta$ to obtain the vector representation $e \in \mathbf{R}^D$ and append it to this sequence. Thus, the new visual token sequence becomes $S^i_v = \left[v^i_1, v^i_2, \dots, v^i_m, e^i_v\right]$, where $v_k^i \in \mathbb{R}^D, k \in \left[1,2,/dots,m\right]$. Similarly, for the caption paired with this image, we also append an  $<$EOS$>$ token to the text token sequence and pass it through the embedding layer of the LLM to obtain the text embedding sequence $S^i_t = \left[t^i_1, t^i_2, \dots, t^i_n, e^i_t\right]$, where $t_k^i \in \mathbb{R}^D,  k \in \left[1,2,/dots,n\right]$. Subsequently, the visual embedding sequence $S_v$ and the text embedding sequence $S_v$ are individually passed through the LLM $\mathbf{L}_\beta$ to obtain the final output from the last layer of $\mathbf{L}_\beta$ as following:
 \begin{small}
\begin{equation}
    H^i_t = \mathbf{L}_\beta\left(S_t^i\right)
\end{equation}
\begin{equation}
    H^i_v = \mathbf{L}_\beta\left(S_v^i\right)
\end{equation}
 \end{small}
where $H^i_v = \left[\hat{v}^i_1, \hat{v}^i_2, \dots, \hat{v}^i_m, \hat{e}^i_v\right]$ and $H^i_t = \left[\hat{t}^i_1, \hat{t}^i_2, \dots, \hat{t}^i_n, \hat{e}^i_t\right]$. Afterwards, we obtain the global representation $\hat{e}^i_v$ that captures the overall semantic information of the image $I_i$, as well as the global representation $\hat{e}^i_t$ that captures the overall semantic information of the ground truth caption $T_i$.


Afterwards, similar to many existing methods in the field of vision-language pretraining \cite{li2021align, li2022blip, bao2022vlmo, yang2022triple, zeng2021multi, jiangbus, jiang2023copa, jiangtrips}, we introduce the following contrastive learning strategy. Assuming a batch size of B during the training process, we compute the text-to-image contrastive learning loss as follows:
\vspace{-2ex}
\begin{small}
\begin{align}
   &  \mathcal{L}_{CL}^t = -\sum\limits_{i=1:B} \frac{1}{B}\mathop{log} 
                                \left[
                                \frac{
                                    f\left(  \hat{e}^i_t, \hat{e}^i_v\right)
                                }
                                {
                                    f\left( \hat{e}^i_t,\hat{e}^i_v\right)+ \sum\limits_{k \neq i} 
                                    f\left( \hat{e}^i_t, \hat{e}^k_v\right)
                                 }
                                \right]
\label{eq1}
\end{align}
\end{small}
where $f\left( \hat{e}^i_t, \hat{e}^i_v \right)$ measures the distance between $\hat{e}^i_t$ and $\hat{e}^i_v$ in a semantic space. Similar, the image-to-text contrastive learning loss as follows:
\vspace{-2ex}
\begin{small}
\begin{align}
   &  \mathcal{L}_{CL}^v = -\sum\limits_{i=1:B} \frac{1}{B}\mathop{log} 
                                \left[
                                \frac{
                                    f\left(  \hat{e}^i_v, \hat{e}^i_t\right)
                                }
                                {
                                    f\left( \hat{e}^i_v,\hat{e}^i_t\right)+ \sum\limits_{k \neq i} 
                                    f\left( \hat{e}^i_v, \hat{e}^k_t\right)
                                 }
                                \right]
\label{eq1}
\end{align}
\end{small}
\vspace{-2ex}
\begin{figure}[t]
    \centering
    \includegraphics[width=1.0\linewidth]{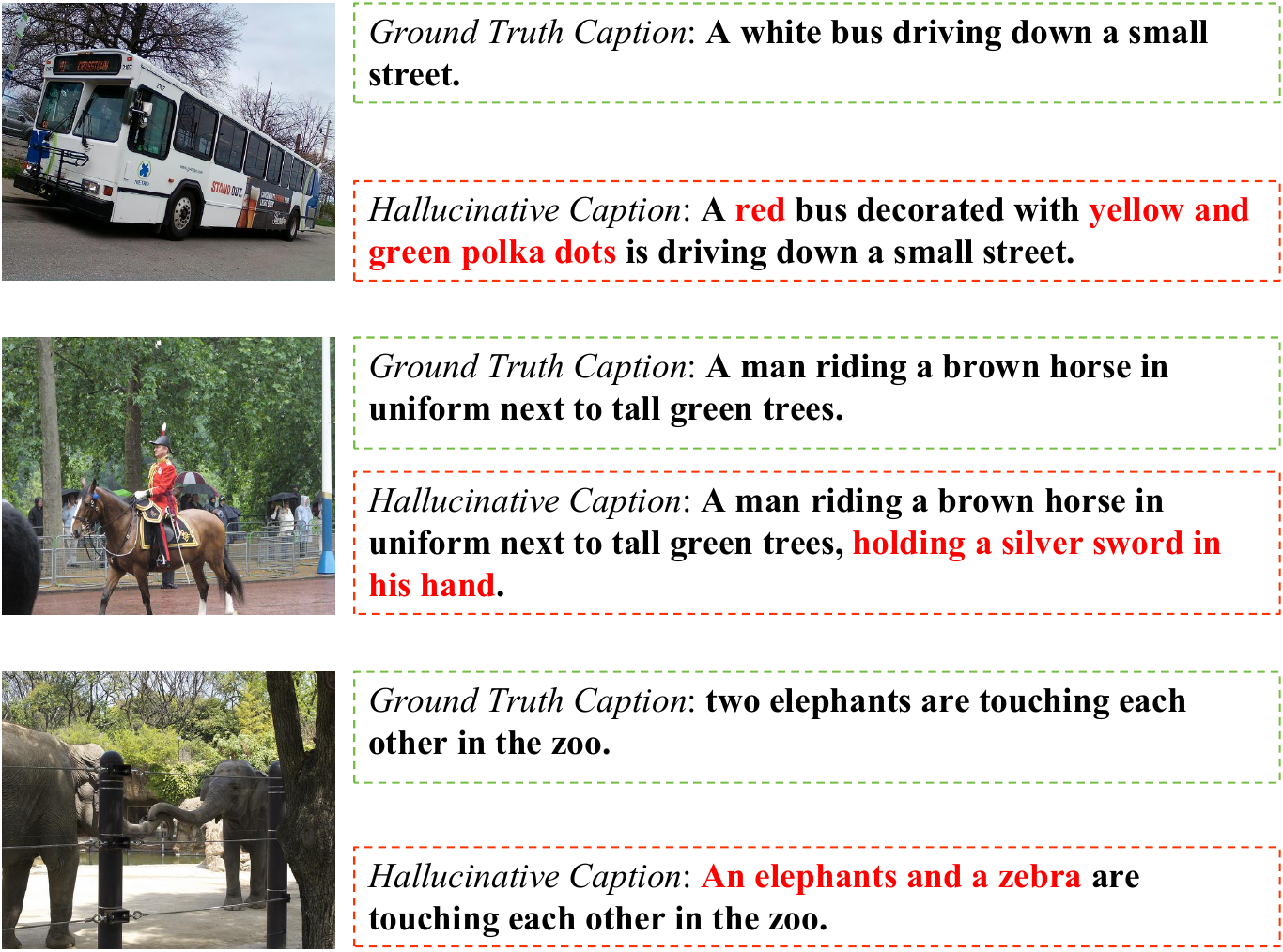}
    \vspace{-2ex}
    \caption{This figure showcases a range of hallucinative captions generated by GPT-4. The hallucinative text is highlighted in red.}
    \label{fig:hallucination_case}
\vspace{-1ex}
\end{figure}
\subsection{Improving Contrastive Learning with Hallucinative Captions}

 We propose to improve the effectiveness of contrastive learning by introducing hard negative samples which mimic the hallucinative text generated by MLLMs. 
 
 \paragraph{Generation of Hallucinative Captions}
 In order to do this, we utilize GPT-4 \cite{OpenAI2023gpt4} to incorporate some elements into the ground truth captions that are either inconsistent with the image content or completely absent from it. As shown in Figure \ref{fig:hallucination_case}, these hallucinations can be coarse-grained, focusing on the presence of objects, or fine-grained, focusing on specific attributes such as quantity, properties, or locations. Here is our prompt to GPT-4: 

 \textit{
 Hallucination in Large-scale Visual Language Models (LVLMs) refers to cases where these models generate descriptions introducing elements that are inconsistent with the content or completely absent from a provided image. These hallucinations can be coarse-grained, focusing on the mere existence of objects, or fine-grained, focusing on more specific attributes or characteristics such as quantity, properties, and locations. 
Your task is to revise a given caption to create a mirrored version that closely aligns with the original's content and length but incorporates elements of hallucination.  The first step involves identifying the objects involved and their associated attributes within the given caption. Subsequently, combine this insight with the details concerning hallucinations provided above to complete your task.
 }
 
 To improve the generation of more appropriate hallucinative captions, we also provide some contextual examples for GPT-4. Please check our appendix for more details.

\paragraph{Hallucination Augmented Contrastive Learning}
 Assuming that we have generated an hallucinative caption $\dot{T}_i$ based on the original caption $T_i$  for the image $I_i$ , and obtained the global representation $\dot{e}^i_t$ of the hallucinative caption using the approach described in subsection \ref{subsec:ILCL}, we can treat it as a negative sample in the image-text contrastive learning. Therefore, the new formula for the image-to-text contrastive learning becomes:
 \vspace{-1ex}
 \begin{small}
\begin{align}
   &  \mathcal{L}_{CL}^v = \nonumber  \\
   &-\sum\limits_{i=1:B+1} \frac{1}{B+1}\mathop{log} 
                                \left[
                                \frac{
                                    f\left(  \hat{e}^i_v, \hat{e}^i_t\right)
                                }
                                {
                                    f\left( \hat{e}^i_v,\hat{e}^i_t\right)+  f\left( \hat{e}^i_v,\dot{e}^i_t\right)+ \sum\limits_{k \neq i} 
                                    f\left( \hat{e}^i_v, \hat{e}^k_t\right)
                                 }
                                \right]
\label{eq1}
\end{align}
\end{small}
For the text-to-image contrastive learning, we have not made changes and have maintained consistency with the content presented in subsection \ref{subsec:ILCL}.
\subsection{Training Paradigm}

As shown in Figure \ref{fig:arch} (b) which demonstrates how \modelname is introduced during the training process of MLLMs. Typically, we incorporate \modelname into the first-stage pretraining of the model to optimize the interface $\mathbf{F}_\alpha$ better. Therefore, suppose the loss function of text generation task is denoted as $\mathcal{L}_G$ and the optimization object of the first stage can be defined as follow:
\begin{small}
    \begin{equation}
    \mathcal{O}_{\alpha}  = \arg\min_{\alpha} \mathcal{L}_G + \left(\mathcal{L}_{CL}^v+  \mathcal{L}_{CL}^t\right)/2
\end{equation}
\end{small}
\noindent In the second stage, we follow the same approach as other methods and fine-tune the model using only instructional data. 

\section{Experiments}
\label{sec:experiments}
\subsection{Implementation}
We validated the effectiveness of our method by applying it to four different models: miniGPT-4 \cite{Zhu2023MiniGPT4}, LLaVA \cite{Liu2023Llava} and LLaVA-1.5 \cite{Liu2023LLava15}.

\begin{table*}[h]
\centering
\resizebox{\textwidth}{!}{%
\begin{tabular}{l|ll|cccccccc}
\hline
\multirow{2}{*}{\textbf{Method}} & \textbf{Overall} & \textbf{Hallucination} & \multicolumn{8}{c}{\textbf{Score in Each Question Type} $\uparrow$} \\
& \textbf{Score} $\uparrow$ & \textbf{Rate} $\downarrow$ & {Attribute} & {Adversarial} & {Comparison} & {Counting} & {Relation} & {Environment} & {Holistic} & {Other} \\
\hline
Kosmos-2 \cite{Peng2023Kosmos2GM} & 1.69 & 0.68 & 2.00 & 0.25 & 1.42 & 1.67 & 1.67 & 2.67 & \textbf{2.50} & 1.33 \\
IDEFICS$_\textsc{9B}$ \cite{laurencon2023idefics} & 1.89 & 0.64 & 1.58 & 0.75 & 2.75 & 1.83 & 1.83 & 2.50 & 2.17 & 1.67 \\
IDEFICS$_\textsc{80B}$ \cite{laurencon2023idefics} & 2.05 & 0.61 & 2.33 & 1.25 & 2.00 & 2.50 & 1.50 & 3.33 & 2.33 & 1.17 \\
InstructBLIP$_\textsc{7B}$ \cite{Dai2023InstructBLIP} & 2.10 & 0.58 & 3.42 & 2.08 & 1.33 & 1.92 & 2.17 & 3.67 & 1.17 & 1.08 \\
InstructBLIP$_\textsc{13B}$ \cite{Dai2023InstructBLIP} & 2.14 & 0.58 & 2.75 & 1.75 & 1.25 & \textbf{2.08} & \textbf{2.50} & \textbf{4.08} & 1.50 & 1.17 \\
LLaVA-RLHF$_\textsc{7B}$ \cite{Sun2023LLavaRlhf} & 2.05 & 0.68 & 2.92 & 1.83 & \textbf{2.42} & 1.92 & 2.25 & 2.25 & 1.75 & 1.08 \\
\hline
LLaVA$_\textsc{7B}$ \cite{Liu2023Llava} & 1.55 & 0.76 & 1.33 & 0.00 & 1.83 & 1.17 & 2.00 & 2.58 & 1.67 &  1.83 \\
 
LLaVA$_\textsc{7B}$-\modelname \cite{Liu2023Llava} & 2.08 ($\uparrow 0.53$) & 0.62 ($\downarrow 0.15$) & 2.94 & 2.01 & 2.27 & 1.64 & 2.35 & 2.14 & 1.67 &  1.63 \\
\hline
miniGPT-4$_\textsc{7B}$ \cite{Liu2023Llava} & 1.39 & 0.71 & 0.75 & 1.83 & 2.16 & 0.91 & 1.25& 1.33& 0.91&  1.91 \\
 
miniGPT-4$_\textsc{7B}$-\modelname          & 1.80 ($\uparrow 0.31$)  & 0.65 ($\downarrow 0.06$)  & 1.22  & 1.85 & 2.23 & 1.74 & 2.13 & 2.48 & 1.03 &   1.58\\
\hline
LLaVA1.5$_\textsc{7B}$ \cite{Liu2023Llava}  & 2.08 & 0.52         & 2.75 & 2.00& 2.33& \textbf{2.08} & 1.50 & 1.91 & 1.91& 2.16 \\
 
LLaVA1.5$_\textsc{7B}$-\modelname   & \textbf{2.13} ($\uparrow 0.05$) & \textbf{0.50}  ($\downarrow 0.02$)  & \textbf{2.95}  & \textbf{2.15} &  2.29 &  1.97 &  1.53 & 1.98 & 2.02 & \textbf{2.19}  \\
\hline
\end{tabular}
}
\vspace{-2ex}
\caption{Evaluation results for different MLLMs on MMHal-Bench. }

\label{tab:mmhalbench_eval}
\end{table*}

\begin{table*}[h!]
\centering
\tiny
 
\begin{tabular}{l|l|cccc|cc|cc|cc}
\toprule
Datasets & Metrics  & Shikra \cite{Chen2023Shikra} & InstructBLIP \cite{Dai2023InstructBLIP} & MM-GPT \cite{gong2023mmgpt} & mPLUG-Owl \cite{ye2023mplugowl} & \multicolumn{2}{c}{MiniGPT-4 \cite{Zhu2023MiniGPT4} w/ \modelname} & \multicolumn{2}{c}{LLaVA \cite{Liu2023Llava} w/ \modelname} &  \multicolumn{2}{c}{LLaVA1.5 \cite{Liu2023LLava15} w/ \modelname}  \\
\cmidrule(lr){1-2}\cmidrule(lr){3-8}\cmidrule(lr){9-10} \cmidrule(lr){11-12} 
\multirow{5}{*}{Random}
& Accuracy ($\uparrow$)       & 86.90 & 88.57 & 50.10  & 53.97 & 54.64 &  80.49 ($\uparrow 25.84$ ) & 50.97 & 82.16 ($\uparrow 31.18$) &  88.17 & 88.59 ($\uparrow 0.42$)   \\
& Precision ($\uparrow$)      & 94.40 & 84.09 & 50.05  & 52.07 & 57.92 &  94.32 ($\uparrow 36.39$)  & 50.19 & 87.30 ($\uparrow 37.11$)&  97.68 & 98.62 ($\uparrow 0.93$   \\
& Recall ($\uparrow$)         & 79.27 & 95.13 & 100.00 & 99.60 & 34.65 &  75.34 ($\uparrow 40.69$)  & 99.13  & 76.53 ($\downarrow 22.59$) &  78.93  & 80.60 ($\uparrow 1.66$) \\
& F1-Score ($\uparrow$)       & 86.19 & 89.27 & 66.71  & 68.39 & 43.35 &  83.82 ($\uparrow 40.46$)  & 66.71 & 81.56 ($\uparrow 14.85$)  &  87.31 & 88.70 ($\uparrow 1.39$)  \\
& Yes ($\rightarrow 50\%$)    & 43.26 & 56.57 & 98.90  & 95.63 & 31.32 &   44.33 ($\uparrow 13.01$) & 99.90 & 45.19 ($\downarrow 54.71$) &   41.64 & 44.43 ($\uparrow 2.78$) \\
\cmidrule(lr){1-2}\cmidrule(lr){3-8}\cmidrule(lr){9-10} \cmidrule(lr){11-12} 
\multirow{5}{*}{Popular}
& Accuracy ($\uparrow$)       & 83.97 & 82.77 & 50.00  & 50.90 & 56.67 & 78.32 ($\uparrow 21.64$) & 49.87 & 79.32 ($\uparrow 29.44$) & 87.46  & 87.94 ($\uparrow 0.48$)\\
& Precision ($\uparrow$)      & 87.55 & 76.27 & 50.00  & 50.46 & 58.69 & 79.23 ($\uparrow 20.54$) & 49.93 & 80.34 ($\uparrow 30.41$) & 95.17  & 97.23 ($\uparrow 2.06$)\\
& Recall ($\uparrow$)         & 79.20 & 95.13 & 100.00 & 99.40 & 44.74 & 74.54 ($\uparrow 29.80$) & 99.27 & 76.60 ($\downarrow 22.67$)& 78.93  & 79.31 ($\uparrow 0.37$)\\
& F1-Score ($\uparrow$)       & 83.16 & 84.66 & 66.67  & 66.94 & 50.74 & 76.85 ($\uparrow 26.11$) & 66.44 & 78.43 ($\uparrow 11.99$)& 86.29  & 87.36 ($\uparrow 1.07$)\\
& Yes ($\rightarrow 50\%$)    & 45.23 & 62.37 & 100.00 & 98.57 & 62.20 & 45.23 ($\downarrow 16.97$) & 99.40 & 47.64  ($\downarrow 51.76$)& 41.46 & 45.03 ($\uparrow 3.57$)\\
\cmidrule(lr){1-2}\cmidrule(lr){3-8}\cmidrule(lr){9-10} \cmidrule(lr){11-12} 
\multirow{5}{*}{Adversarial}
& Accuracy ($\uparrow$)       & 83.10 & 72.10 & 50.00  & 50.67 & 54.50 & 71.32 ($\uparrow 16.82$)&  49.70& 74.47  ($\uparrow 24.77$)& 85.93 & 86.54 ($\uparrow 0.61$)\\
& Precision ($\uparrow$)      & 85.60 & 65.13 & 50.00  & 50.34 & 57.21 & 70.53 ($\uparrow 13.32$)& 49.85 & 73.55  ($\uparrow 23.70$)& 91.78 & 93.01 ($\uparrow 1.23$)\\
& Recall ($\uparrow$)         & 79.60 & 95.13 & 100.00 & 99.33 & 41.45 & 73.45 ($\uparrow 32.00$)& 99.07 & 76.40  ($\downarrow 22.67 $)& 78.93 &  79.52 ($\uparrow 0.59$)\\
& F1-Score ($\uparrow$)       & 82.49 & 77.32 & 66.67  & 66.82 & 48.07 & 71.96 ($\uparrow 23.89$)& 66.32 & 74.95  ($\uparrow 8.63$)& 84.87 &  85.73 ($\uparrow 0.86$)\\
& Yes ($\rightarrow 50\%$)    & 46.50 & 73.03 & 100.00 & 98.67 & 38.32 & 48.23 ($\uparrow 9.91$)& 99.37 & 51.93  ($\downarrow 47.44 $)& 43.00 &  46.33 ($\uparrow 3.33$)\\
\bottomrule
\end{tabular}

\caption{Object hallucination benchmark using POPE \cite{Li2023pope} evaluation pipeline . "Yes" signifies the likelihood of the model producing a positive response.
}
\label{table:pope_result}
\end{table*}
\paragraph{Data sets} 

For MiniGPT-4, the pre-training phase utilized significantly large datasets such as LAION\cite{schuhmann2022laion} (115 million), Conceptual Captions \cite{changpinyo2021cc3m12m} (CC3M/CC12M), and others. However, generating hallucinative captions for such enormous datasets is very costly. As a result, for MiniGPT-4, we randomly sampled about 10 million data, representing 10\% of the total, and didn't use hallucinative captions for contrastive learning for the remaining data during training.  Moreover, we discovered that regardless of not using hallucinative captions for enhancement, our model still significantly enhances models such as MiniGPT-4 \cite{Zhu2023MiniGPT4}. On the other hand, for the LLaVA \cite{Liu2023Llava} and LLaVA1.5 \cite{Liu2023LLava15}, which used subsets of LAION/CC/SBU datasets with roughly 558K data, we generated hallucinative captions for every training datum.

\paragraph{Training Settings}
We followed the original approach for MiniGPT-4 \cite{Zhu2023MiniGPT4} and retrained it using the complete pre-training dataset, about 10M data included hallucinative captions. For LLaVA \cite{Liu2023Llava} and LLaVA 1.5 \cite{Liu2023LLava15}, we used the complete pre-training dataset introduced \modelname during the first stage of pre-training.  We keeping the same hyperparameter settings for all above models. Our experiments were conducted using 16 NVIDIA A100 GPUs with 80G of memory. Due to the increased memory usage during MLLMs training (which includes model and gradient data), the batch size during contrastive learning was affected. To address this, we used a queue of size 16,384, similar to the approaches used for ALBEF \cite{li2021align} and MOCO \cite{chen2020moco}, to store more negative samples.  We used Deepspeed \cite{rajbhandari2020zero} for LLaVA and LLaVA 1.5, with a batch size of 64 and 32 on a single GPU, respectively. For MiniGPT-4, the batch size was 8.

\subsection{Effectiveness of \modelname on Mitigating Hallucination}

To verify the efficacy of our proposed method in addressing hallucination issues, we leveraged two widely used benchmark evaluation datasets that evaluate the presence of hallucinations in models. These datasets included MMHal-Bench \cite{Sun2023LLavaRlhf} and POPE \cite{Li2023pope}. MMHal-Bench offers a comprehensive evaluation of models that encompasses multiple perspectives, such as attributes, relations, and counting. On the other hand, POPE particularly focuses on hallucinations related to objects. We employed both datasets to measure the effectiveness of our method in addressing hallucination across various scenarios.

\paragraph{Evaluation on MMHal-Bench}
For the MMHal-Bench \cite{Sun2023LLavaRlhf}.  We apply our method to iniGPT-4 \cite{Zhu2023MiniGPT4}, LLaVA \cite{Liu2023Llava}, LLaVA1.5 \cite{Liu2023LLava15}  and  compare the results with other recent vision-language models, including MKosmos-2 \cite{Peng2023Kosmos2GM}, IDEFICS \cite{laurencon2023idefics}, InstructBLIP \cite{Dai2023InstructBLIP}, and anther LLaVA-RLHF \cite{Sun2023LLavaRlhf}. Following \cite{Sun2023LLavaRlhf}, we use GPT-4 to evaluate the overall score and hallucination rate of different MLLMs. Table \ref{tab:mmhalbench_eval} demonstrates a significant improvement in the overall performance of MMHal-Bench after applying our method to LLaVA \cite{Liu2023Llava}, MiniGPT-4\cite{Zhu2023MiniGPT4}, and LLaVA1.5\cite{Liu2023LLava15}. Notably, MiniGPT-4-\modelname exhibited considerable performance gain over MiniGPT-4 \cite{Zhu2023MiniGPT4}.  Moreover, compared with LLaVA-RLHF\cite{Sun2023LLavaRlhf}, a recently proposed method that uses human feedback and reinforcement learning to address hallucinations,  LLaVA-\modelname showed an even more significant improvement. 


\paragraph{Evaluation on POPE} In addition, we obtained consistent results using MMHal-Bench \cite{Sun2023LLavaRlhf} in the POPE evaluation benchmark \cite{Li2023pope}. Table \ref{table:pope_result} shows that miniGPT-4-\modelname and LLaVA-\modelname both demonstrated significant improvements compared to the original model. Of particular note, the average F1 score of LLaVA-\modelname increased by 17.8\% compared to LLaVA \cite{Liu2023Llava}, while the Yes ratio decreased from 99.55 to 48.25. Furthermore, by applying our method to LLaVA1.5 \cite{Liu2023LLava15}, LLaVA1.5-\modelname easily achieved SOTA on this benchmark. Noted that LLaVA1.5 \cite{Liu2023LLava15} is a high-performing model with a low likelihood of generate hallucination, surpassing MiniGPT-4 \cite{Zhu2023MiniGPT4} and LLaVA \cite{Liu2023Llava}. This model's impressive benchmark scores make it a valuable foundation to build upon.




\subsection{Effectiveness of \modelname on Visual Comprehension}

\modelname has shown effectiveness in solving the issue of hallucination. Nevertheless, we intend to explore the influence of \modelname on the model's abilities of visual comprehension and generation. To achieve this objective, we carried out assessments on common benchmarks, such as Visual Question Answering (VQA) \cite{balanced_vqa_v2,mishra2019ocrvqa, singh2019textvqa} after incorporating \modelname into the MLLMs. Furthermore, as MLLMs possess robust zero-shot capabilities, traditional evaluation metrics often fail to provide a detailed assessment of their abilities. Additionally, their inability to match the given answer correctly exacerbates significant robustness issues. To mitigate these challenges, the research community introduced a series of benchmarks. These benchmarks aim to systematically structure and evaluate complex multi-modal tasks from various perspectives. Therefore, we also evaluated the model's performance on recently designed MLLM-focused Multi-modal Benchmarks including MME \cite{fu2023mme}, MMBench \cite{liu2023mmbench}, MM-Vet \cite{yu2023mmvet}, SEED-Bench \cite{li2023seedbench}.

\begin{table*}
\centering
    
    \begin{tabular}{l|c|cc|ccc}
        \shline
         ~ & ~  & \multicolumn{2}{c}{General VQA} & \multicolumn{3}{c}{General VQA (Zero-shot)}  \\
         \cmidrule(lr){3-4} \cmidrule(lr){5-7}
        \multirow{2}{*}{Method} & \multirow{2}{*}{\#Params} & \multirow{2}{*}{VQAv2} & \multirow{2}{*}{GQA} & \multirow{2}{*}{VizWizQA} & \multirow{2}{*}{TextVQA} & \multirow{2}{*}{SciQA (IMG)} \\
       ~ & ~ & ~ & ~ & ~ & ~ & ~ \\
        \hline
        BLIP-2 \cite{Li2023BLIP2} & 8.2B &   65.0 &   41.0 & 19.6 & 42.5 & 61.0 \\
         InstructBLIP \cite{Dai2023InstructBLIP} & 8.2B &   - & 49.2 & 34.5 & $50.1^\dag$ & 60.5 \\
        Unified-IO$_{\text{XL}}$ \cite{Lu2022UnifiedIO} & 2.9B &   77.9 &   - & \demph{$57.4^\ddag$} & - & - \\
        PaLM-E-12B \cite{Driess2023PaLME} & 12B &  76.2 &   - & - & - & - \\
        Shikra \cite{Chen2023Shikra} & 7.2B &  77.4 &  - & - & - & - \\
       
        Qwen-VL-Chat \cite{Bai2023QwenVL} & 9.6B & 78.2 &  57.5 & 38.9 & \demph{$61.5^\ddag$} & \textbf{68.2} \\ \hline
        LLaVA \cite{Liu2023LLava15} & 7.2B & 71.3   &  41.3&  36.7 &  $50.2^\dag$ & 61.5 \\
         
        LLaVA-\modelname \cite{Liu2023LLava15} & 7.2B & 73.3  & 42.5 & 37.4 & $52.2^\dag$ & 62.4 \\ \hline
        MiniGPT-4 \cite{Liu2023LLava15} & 7.2B & 65.2   & 30.8 & 30.2 & $52.3^\dag$ & 58.4 \\
         
        MiniGPT-4-\modelname \cite{Liu2023LLava15} & 7.2B & 68.9 & 32.3& 31.7 & $54.2^\dag$ & 60.3\\ \hline
        LLaVA1.5 \cite{Liu2023LLava15} & 7.2B & 78.5  & 62.0 & 50.0 & $58.2^\dag$ & 66.8 \\
         
        LLaVA1.5-\modelname \cite{Liu2023LLava15} & 7.2B & \textbf{79.1} & \textbf{62.5} & \textbf{50.5} & $59.8^\dag$ & 67.3 \\
        \shline
    \end{tabular}

    \caption{\textbf{Performance comparison on visual question answering.} For VQA, accuracy is reported. Note that specialists are fine-tuned on each individual dataset. \dag\ denotes OCR inputs are utilized. \ddag\ indicates the model has trained on the dataset. We gray out those specialists' methods which are individually fine-tuned on the dataset as well as those fine-tuned results of generalists.
    }
 
    \label{table:multimodal-results}
\end{table*}

\begin{table*}
\centering
   
    \begin{tabular}{l|c|c|c|c|c|c}
        \shline
        Method    & Vision Encoder & Language Model & MME     & MMBench & MM-Vet & SEED-Bench  \\
        \hline
        BLIP-2 \cite{Li2023BLIP2}      & ViT-g (1.3B)            & Vicuna (7B)             & 1293.84 & -       & 22.4   & 46.4       \\
        mPLUG-Owl \cite{ye2023mplugowl} & ViT-L (0.3B)    & LLaMA (7B) & 967.34 & 46.6 & - & 34.0 \\
        InstructBLIP \cite{Dai2023InstructBLIP}  & ViT-g (1.3B)            & Vicuna (7B)             & 1212.82 & 36.0    & 26.2   & 53.4     \\
        LLaMA-Adapter-v2 \cite{Gao2023LLaMAAdapterV2} & ViT-L (0.3B)            & LLaMA (7B)              & 1328.40 & 39.5    & 31.4   & 32.7      \\
        Otter \cite{Li2023Otter}           & ViT-L (0.3B)            & LLaMA (7B)              & 1292.26 & 48.3    & 24.6   & 32.9    \\
        Qwen-VL-Chat \cite{Bai2023QwenVL}    & ViT-G (1.9B)            & Qwen (7B)               & 1487.58 & 60.6    & -      & 58.2   \\ \hline
         LLaVA \cite{Liu2023Llava}      & ViT-L (0.3B)             & Vicuna (7B)             & 502.82  & 36.2    & 28.1   & 33.5       \\
          
          LLaVA-\modelname \cite{Liu2023Llava}      & ViT-L (0.3B)            & Vicuna (7B)             & 562.58  & 37.8    & 28.4   & 33.9       \\\hline
         MiniGPT-4 \cite{Zhu2023MiniGPT4}  & ViT-g (1.3B)          & Vicuna (7B)             & 581.67  & 23.0    & 22.1   & 42.8        \\
          
          MiniGPT-4-\modelname  \cite{Zhu2023MiniGPT4}  & ViT-g (1.3B)            & Vicuna (7B)             & 653.94  & 24.5    & 23.8   & 42.5        \\\hline
        LLaVA-1.5 \cite{Liu2023LLava15}     & ViT-L (0.3B)            & Vicuna (7B)             & 1510.70 & 64.3    & \textbf{30.5}   &  58.6   \\    
         
         LLaVA-1.5-\modelname  \cite{Liu2023LLava15}     & ViT-L (0.3B)            & Vicuna (7B)             & \textbf{1530.10} & \textbf{64.5}   & 30.4   &  \textbf{58.9}   \\   
        \shline
    \end{tabular}

    \caption{\textbf{Zero-shot multi-modal evaluation on multi-modal benchmarks} including MME \cite{fu2023mme}, MMBench \cite{liu2023mmbench}, MM-Vet \cite{yu2023mmvet}, SEED-Bench \cite{li2023seedbench}. The overall scores are reported for evaluation. For MMBench, we report test results.}
 \vspace{-1ex}
    \label{table:zeroshot-multimodal-bench}
\end{table*}

\paragraph{Results on Benchmark Tasks}
 Our evaluation includes six popular benchmarks, as summarized in Table \ref{table:multimodal-results}. 
We applied the \modelname to three baselins: MiniGPT-4, LLaVA, and LLaVA1.5, and compared their performance to other State-of-the-Art (SOTA) MLLMs such as BLIP2\cite{Li2023BLIP2}, InstructBLIP \cite{Dai2023InstructBLIP}, Shikra \cite{Chen2023Shikra}, and Qwen-VL-Chat \cite{Bai2023QwenVL}. Our experimental results show that our approach successfully enhances the performance of original models across a range of VQA datasets. Notably, LLaVA-\modelname outperforms LLaVA \cite{Liu2023Llava} in terms of consistency and accuracy across all VQA datasets. Additionally, when compared to LLaVA1.5 \cite{Liu2023LLava15}, LLaVA1.5-\modelname achieves better results in General VQA benchmarks abd zero-shot VQA tasks, implying that MLLMs may not only mitigate hallucinations but also improve correlations between visual and textual information, which further refines the generalization ability of models.


\paragraph{MLLM-oriented Multi-modal Benchmarks.} 

 We applied \modelname to MiniGPT-4 \cite{Zhu2023MiniGPT4}, LLaVA \cite{Liu2023Llava}, LLaVA1.5 \cite{Liu2023LLava15} and evaluate them on five recently popular multi-modal benchmarks  in a zero-shot manner. For a fair comparison, we select models with similar language model sizes, particularly those from the LLaMA \cite{Touvron2023LLaMA} family, and detail their differences in the vision encoder. 
 The results of our evaluation are listed in Table \ref{table:zeroshot-multimodal-bench}. We discovered that after implementing \modelname, all three models exhibited improvements across multiple benchmarks.  Notably, for LLaVA and MiniGPT-4, the enhancement was particularly evident on the MME \cite{fu2023mme} benchmark. For instance, after implementing \modelname, LLaVA's MME score improved from 581.67 to 653.94. These results indicate that our methodology can not only reduce the instances of model hallucination but also enhance the model's visual comprehension capabilities. 
 
 \begin{figure*}[t]
  \centering
  \begin{subfigure}[b]{0.3\linewidth}
    \includegraphics[width=\textwidth]{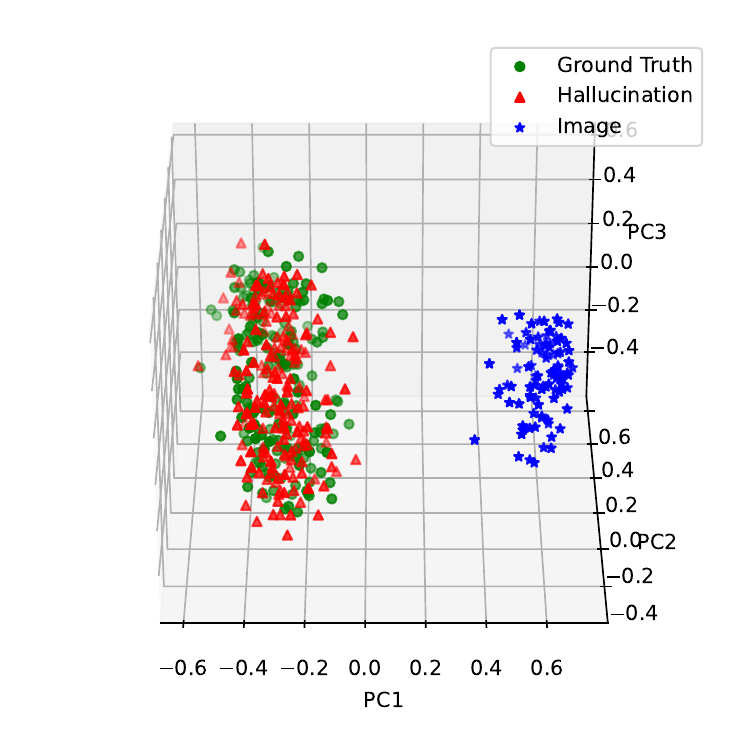}
     \vspace{-3ex}
    \caption{w/o \modelname}
    \label{fig:sub1}
  \end{subfigure}
  \hfill
  \begin{subfigure}[b]{0.3\linewidth}
    \includegraphics[width=\textwidth]{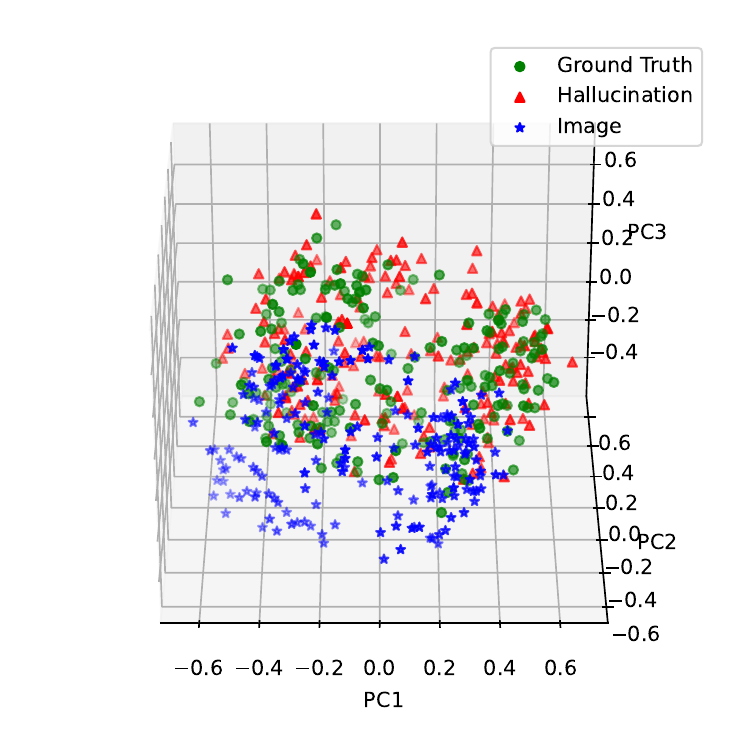}
     \vspace{-3ex}
    \caption{w/ CL }
    \label{fig:sub2}
  \end{subfigure}
  \hfill
  \begin{subfigure}[b]{0.3\linewidth}
    \includegraphics[width=\textwidth]{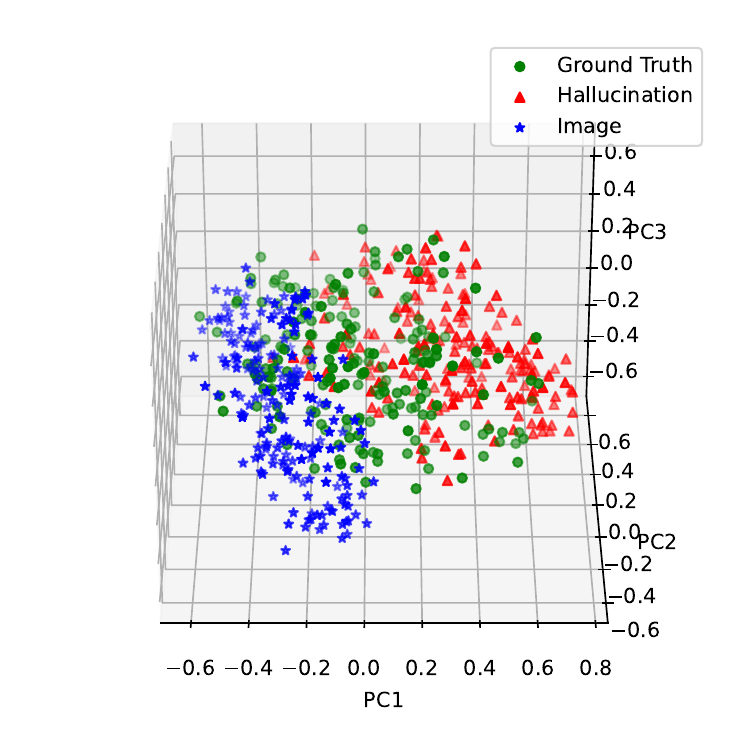}
    \vspace{-3ex}
    \caption{w/ \modelname}
    \label{fig:sub3}
  \end{subfigure}
  \vspace{-2ex}
  \caption{This figure illustrates the visualization of various data distributions. The blue icons represent visual data extracted from images, the green icons denote ground truth caption data, and the red icons signify hallucinative caption data. The label "w/o \modelname" represents the data distribution obtained from the original model output without employing our proposed method. On the other hand, "w/ CL" indicates the data distribution resulting from the model output utilizing contrastive learning. Lastly, "w/ \modelname" indicates the data distribution generated by the model output using our proposed method.}
  \vspace{-1ex}
  \label{fig:three_images}
\end{figure*}
\subsection{Ablation Study}
\begin{table}[htbp]
\footnotesize
 \setlength{\tabcolsep}{1.7mm}{
\begin{tabular}{lcccccc}
\toprule[1.0pt]
Model & w/ CL & w/ HC & POPE & MMHal & VQA & MME   \\ \hline
LLaVA   & $\times$  & $\times$     &    66.48   &  1.55   &   71.32  & 502.82 \\
LLaVA   & \checkmark &  $\times$    &   69.23   &  1.67   &   72.98  &  549.04 \\
LLaVA   & \checkmark  & \checkmark  &   78.31   & 2.08    &  73.30   & 562.58  \\
\hline
MiniGP-4 &   $\times$  & $\times$      &  47.38  &   1.39  & 65.2  & 581.67  \\
MiniGP-4   &\checkmark &  $\times$     &  53.54  &  1.45   & 67.6 & 633.21  \\ 
MiniGP-4   &\checkmark  & \checkmark   &  77.54  &  1.80  & 68.9  & 653.94  \\  
\hline
LLaVA1.5  & $\times$  & $\times$      &   86.15  & 2.08   &   78.5  &  1510.70 \\ 
LLaVA1.5   &\checkmark &  $\times$    &   86.31  & 2.09   &  78.7 & 1523.84  \\ 
LLaVA1.5  &\checkmark  & \checkmark   &   87.26  &  2.13   &   79.1 & 1530.10 \\ \bottomrule[1.0pt]
\end{tabular}}
\centering
\vspace{-2ex}
\caption{The result of ablations for the impact of hallucinative captions.  We report the text-dev score results of POPE\cite{Li2023pope}, MMhal-Bench \cite{Sun2023LLavaRlhf}, VQA and MME.  w/ CL refers to training MLLMs with Contrastive Learning for MLLMs, w/ HC refers to utilize hallucinative captions to enhance the contrastive learning. For POPE, we report the average F1 score. }
\vspace{-2ex}
\label{table:ablation1}
\end{table}

\begin{table}[htbp]
\centering
\footnotesize
\setlength{\tabcolsep}{1.3mm}{
\begin{tabular}{@{}lcccccc@{}}
\toprule[1.0pt]
model  & +VE & +LLM &  POPE & MMHal & VQA & MME \\ \midrule
LLaVA-\modelname &  \checkmark &  $\times$    & 63.42  & 1.43 & 65.0 & 324.50 \\ 
LLaVA-\modelname &  $\times$  &  \checkmark & 78.53  & 2.08 & 74.2 & 580.32 \\ 
LLaVA-\modelname &  $\times$  & $\times$  & 78.31   & 2.08    &  73.3  & 562.58  \\ \hline
LLaVA1.5-\modelname&   \checkmark &  $\times$     & 68.89  &  1.53  &  69.6   & 459.34 \\ 
LLaVA1.5-\modelname & $\times$  &  \checkmark & 87.23  &  2.13   &  79.4 & 1542.48  \\ 
LLaVA1.5-\modelname &  $\times$  & $\times$  & 87.26  &  2.13   &  79.1  & 1530.10 \\ 
\bottomrule[1.0pt]
\end{tabular}}
\vspace{-2ex}
\caption{Results of models under different training paradigms. "+VE" denotes training the Visual Encoder during Stage 1 pretraining, while "+LLM" indicates training the LLM during Stage 1 pretraining. }
\label{table6}
\vspace{-2ex}
\end{table}
\paragraph{Impact of Hallucinative Captions}
To validate the effectiveness of using hallucinative captions as hard negative samples in contrastive learning for resolving hallucinations, we conducted the following experiments: In the Stage 1 pre-training phase, we did not introduce any additional hallucinative captions, and the contrastive learning loss was calculated solely based on the equations 3 and 9 discussed in subsection 3.1 of our paper. We conducted experiments on MLLMs including LLaVA \cite{Liu2023Llava}, MiniGPT-4 \cite{Zhu2023MiniGPT4}, and LLaVA1.5 \cite{Liu2023LLava15}, and reported the results on benchmarks such as POPE and MMHal-Bench. Additionally, we also reported results on the MME and VQA benchmarks.
As illustrated in the Table \ref{table:ablation1}, absent the facilitation from hallucinative captions, the models displayed moderate improvements on hallucination benchmarks such as MMhal-Bench, yet these improvements were somewhat constrained. However, the subsequent inclusion of hallucinative captions resulted in a marked enhancement on the same hallucination benchmark, thus affirming the potency of the hallucinative captions. Furthermore, we observed analogous improvements in the model's performance on both MME and VQA. Our hypothesis asserts that hallucinative captions aid MLLMs in diverting the visual representation from hallucinations and other textual inaccuracies. This action helps avoid instances of hallucination. Furthermore, contrastive learning supports the model by aligning the semantics of image-text, which ultimately enhances the model's effectiveness.

\paragraph{Discussion on Training Paradigm}
We have observed that certain Multimodal Language-and-Vision Models (MLLMs) may not freeze the activity of either the visual encoder or the Language-and-Vision Models (LLMs) during the initial stage of pretraining. To assess the impact of our methodology under such distinct training paradigms, we independently tested models where either the Visual Encoder or the LLMs were active during the first pretraining phase. These tests were conducted on two platforms: LLaVA and LLaVA1.5 and subsequently evaluated against multiple benchmark standards.  As illustrated in Table \ref{table6}, the models experienced a significant performance decline when LLMs are activated. We hypothesize that this downturn could be linked to low-quality data in the first pretraining stage and the introduction of additional contrast learning tasks, both of which affect the LLMs' representation distribution. This culminates in the catastrophic forgetting of the LLMs. Conversely, initiating the Visual Encoder led to a modest performance boost. This might be attributed to the fact that the target parameters our model can optimize extend beyond the learnable interface and incorporate the visual encoder as well. This expanded scope paves the way for a more successful alignment of visual and text representations within the MLLMs.

\subsection{Visualization}
The objective of our research is the introduction of \modelname, to further enhance the visual representation output of our interface. The aim is to closely align the output to the correct textual representation within the representation space of Language Models (LLMs) and, at the same time, distance it from hallucinative and other incorrect textual representations.
To substantiate our objective, we randomly selected 200 image-text pairs from the COCO \cite{lin2014coco} val2017 dataset. Using GPT-4, we generated hallucination samples and subsequently reduced these samples using the hidden state representation of the last token through LLMs for visualization purposes. The data distribution under three conditions: without employing \modelname, instigating cross-modal contrastive learning but without the use of hallucination-enhanced samples, and usage of hallucination-enhanced sample contrast learning was visualized respectively. The MLLM utilized in our study was LLaVA.
As illustrated in Figure \ref{fig:three_images} (a), a substantial modality gap is observable in the data distribution without contrast learning. In Figure \ref{fig:three_images} (b), after applying contrast learning, although the modal gap decreased, a differentiation in the distribution of hallucination samples and ground truth samples was unattainable. In Figure \ref{fig:three_images} (c), with the application of hallucination augmentation in contrast learning, not only did the modal gap decrease, but the hallucination sample distribution was also significantly distanced.

\label{subsec:visualization}

\section{Conclusion}
This paper addresses the issue of hallucinations in Multi-modal Large Language Models (MLLMs) and proposes a method called Hallucination Augmented Contrastive Learning (HACL) to improve the alignment between visual and textual representations. By using contrastive learning on projected text and visual token sequences, and incorporating hallucinative captions as hard negative samples, HACL effectively reduces the occurrence of hallucinations. Experimental results demonstrate that incorporating HACL enhances the performance of MLLMs and significantly reduces the occurrence of hallucinations in benchmark evaluations.
{
    \small
    \bibliographystyle{ieeenat_fullname}
    \bibliography{main}
}


\end{document}

%% file: preamble.tex
\usepackage[dvipsnames]{xcolor}


%% file: main.bbl
\begin{thebibliography}{55}
\providecommand{\natexlab}[1]{#1}
\providecommand{\url}[1]{\texttt{#1}}
\expandafter\ifx\csname urlstyle\endcsname\relax
  \providecommand{\doi}[1]{doi: #1}\else
  \providecommand{\doi}{doi: \begingroup \urlstyle{rm}\Url}\fi

\bibitem[Alayrac et~al.(2022)Alayrac, Donahue, Luc, Miech, Barr, Hasson, Lenc, Mensch, Millican, Reynolds, et~al.]{alayrac2022flamingo}
Jean-Baptiste Alayrac, Jeff Donahue, Pauline Luc, Antoine Miech, Iain Barr, Yana Hasson, Karel Lenc, Arthur Mensch, Katherine Millican, Malcolm Reynolds, et~al.
\newblock Flamingo: a visual language model for few-shot learning.
\newblock \emph{Advances in Neural Information Processing Systems}, 35:\penalty0 23716--23736, 2022.

\bibitem[Bai et~al.(2023)Bai, Bai, Yang, Wang, Tan, Wang, Lin, Zhou, and Zhou]{Bai2023QwenVL}
Jinze Bai, Shuai Bai, Shusheng Yang, Shijie Wang, Sinan Tan, Peng Wang, Junyang Lin, Chang Zhou, and Jingren Zhou.
\newblock Qwen-vl: A frontier large vision-language model with versatile abilities.
\newblock \emph{ArXiv}, abs/2308.12966, 2023.

\bibitem[Bao et~al.(2022)Bao, Wang, Dong, Liu, Mohammed, Aggarwal, Som, Piao, and Wei]{bao2022vlmo}
Hangbo Bao, Wenhui Wang, Li Dong, Qiang Liu, Owais~Khan Mohammed, Kriti Aggarwal, Subhojit Som, Songhao Piao, and Furu Wei.
\newblock Vlmo: Unified vision-language pre-training with mixture-of-modality-experts.
\newblock \emph{Advances in Neural Information Processing Systems}, 35:\penalty0 32897--32912, 2022.

\bibitem[Brown et~al.(2020)Brown, Mann, Ryder, Subbiah, Kaplan, Dhariwal, Neelakantan, Shyam, Sastry, Askell, Agarwal, Herbert-Voss, Krueger, Henighan, Child, Ramesh, Ziegler, Wu, Winter, Hesse, Chen, Sigler, Litwin, Gray, Chess, Clark, Berner, McCandlish, Radford, Sutskever, and Amodei]{Brown2020gpt3}
Tom~B. Brown, Benjamin Mann, Nick Ryder, Melanie Subbiah, Jared Kaplan, Prafulla Dhariwal, Arvind Neelakantan, Pranav Shyam, Girish Sastry, Amanda Askell, Sandhini Agarwal, Ariel Herbert-Voss, Gretchen Krueger, T.~J. Henighan, Rewon Child, Aditya Ramesh, Daniel~M. Ziegler, Jeff Wu, Clemens Winter, Christopher Hesse, Mark Chen, Eric Sigler, Mateusz Litwin, Scott Gray, Benjamin Chess, Jack Clark, Christopher Berner, Sam McCandlish, Alec Radford, Ilya Sutskever, and Dario Amodei.
\newblock Language models are few-shot learners.
\newblock \emph{ArXiv}, abs/2005.14165, 2020.

\bibitem[Carion et~al.(2020)Carion, Massa, Synnaeve, Usunier, Kirillov, and Zagoruyko]{carion2020detr}
Nicolas Carion, Francisco Massa, Gabriel Synnaeve, Nicolas Usunier, Alexander Kirillov, and Sergey Zagoruyko.
\newblock End-to-end object detection with transformers.
\newblock In \emph{European conference on computer vision}, pages 213--229. Springer, 2020.

\bibitem[Changpinyo et~al.(2021)Changpinyo, Sharma, Ding, and Soricut]{changpinyo2021cc3m12m}
Soravit Changpinyo, Piyush Sharma, Nan Ding, and Radu Soricut.
\newblock Conceptual 12m: Pushing web-scale image-text pre-training to recognize long-tail visual concepts.
\newblock In \emph{Proceedings of the IEEE/CVF Conference on Computer Vision and Pattern Recognition}, pages 3558--3568, 2021.

\bibitem[Chen et~al.(2023)Chen, Zhang, Zeng, Zhang, Zhu, and Zhao]{Chen2023Shikra}
Ke Chen, Zhao Zhang, Weili Zeng, Richong Zhang, Feng Zhu, and Rui Zhao.
\newblock Shikra: Unleashing multimodal llm's referential dialogue magic.
\newblock \emph{ArXiv}, abs/2306.15195, 2023.

\bibitem[Chen et~al.(2020)Chen, Fan, Girshick, and He]{chen2020moco}
Xinlei Chen, Haoqi Fan, Ross Girshick, and Kaiming He.
\newblock Improved baselines with momentum contrastive learning.
\newblock \emph{arXiv preprint arXiv:2003.04297}, 2020.

\bibitem[Chowdhery et~al.(2022)Chowdhery, Narang, Devlin, Bosma, Mishra, Roberts, Barham, Chung, Sutton, Gehrmann, Schuh, Shi, Tsvyashchenko, Maynez, Rao, Barnes, Tay, Shazeer, Prabhakaran, Reif, Du, Hutchinson, Pope, Bradbury, Austin, Isard, Gur-Ari, Yin, Duke, Levskaya, Ghemawat, Dev, Michalewski, Garc{\'i}a, Misra, Robinson, Fedus, Zhou, Ippolito, Luan, Lim, Zoph, Spiridonov, Sepassi, Dohan, Agrawal, Omernick, Dai, Pillai, Pellat, Lewkowycz, Moreira, Child, Polozov, Lee, Zhou, Wang, Saeta, D{\'i}az, Firat, Catasta, Wei, Meier-Hellstern, Eck, Dean, Petrov, and Fiedel]{Chowdhery2022PaLM}
Aakanksha Chowdhery, Sharan Narang, Jacob Devlin, Maarten Bosma, Gaurav Mishra, Adam Roberts, Paul Barham, Hyung~Won Chung, Charles Sutton, Sebastian Gehrmann, Parker Schuh, Kensen Shi, Sasha Tsvyashchenko, Joshua Maynez, Abhishek Rao, Parker Barnes, Yi Tay, Noam~M. Shazeer, Vinodkumar Prabhakaran, Emily Reif, Nan Du, Benton~C. Hutchinson, Reiner Pope, James Bradbury, Jacob Austin, Michael Isard, Guy Gur-Ari, Pengcheng Yin, Toju Duke, Anselm Levskaya, Sanjay Ghemawat, Sunipa Dev, Henryk Michalewski, Xavier Garc{\'i}a, Vedant Misra, Kevin Robinson, Liam Fedus, Denny Zhou, Daphne Ippolito, David Luan, Hyeontaek Lim, Barret Zoph, Alexander Spiridonov, Ryan Sepassi, David Dohan, Shivani Agrawal, Mark Omernick, Andrew~M. Dai, Thanumalayan~Sankaranarayana Pillai, Marie Pellat, Aitor Lewkowycz, Erica Moreira, Rewon Child, Oleksandr Polozov, Katherine Lee, Zongwei Zhou, Xuezhi Wang, Brennan Saeta, Mark D{\'i}az, Orhan Firat, Michele Catasta, Jason Wei, Kathleen~S. Meier-Hellstern, Douglas Eck, Jeff Dean, Slav Petrov,
  and Noah Fiedel.
\newblock Palm: Scaling language modeling with pathways.
\newblock \emph{J. Mach. Learn. Res.}, 24:\penalty0 240:1--240:113, 2022.

\bibitem[Dai et~al.(2023)Dai, Li, Li, Tiong, Zhao, Wang, Li, Fung, and Hoi]{Dai2023InstructBLIP}
Wenliang Dai, Junnan Li, Dongxu Li, Anthony Meng~Huat Tiong, Junqi Zhao, Weisheng Wang, Boyang~Albert Li, Pascale Fung, and Steven C.~H. Hoi.
\newblock Instructblip: Towards general-purpose vision-language models with instruction tuning.
\newblock \emph{ArXiv}, abs/2305.06500, 2023.

\bibitem[Driess et~al.(2023)Driess, Xia, Sajjadi, Lynch, Chowdhery, Ichter, Wahid, Tompson, Vuong, Yu, Huang, Chebotar, Sermanet, Duckworth, Levine, Vanhoucke, Hausman, Toussaint, Greff, Zeng, Mordatch, and Florence]{Driess2023PaLME}
Danny Driess, F. Xia, Mehdi S.~M. Sajjadi, Corey Lynch, Aakanksha Chowdhery, Brian Ichter, Ayzaan Wahid, Jonathan Tompson, Quan~Ho Vuong, Tianhe Yu, Wenlong Huang, Yevgen Chebotar, Pierre Sermanet, Daniel Duckworth, Sergey Levine, Vincent Vanhoucke, Karol Hausman, Marc Toussaint, Klaus Greff, Andy Zeng, Igor Mordatch, and Peter~R. Florence.
\newblock Palm-e: An embodied multimodal language model.
\newblock In \emph{International Conference on Machine Learning}, 2023.

\bibitem[Fu et~al.(2023)Fu, Chen, Shen, Qin, Zhang, Lin, Qiu, Lin, Yang, Zheng, et~al.]{fu2023mme}
Chaoyou Fu, Peixian Chen, Yunhang Shen, Yulei Qin, Mengdan Zhang, Xu Lin, Zhenyu Qiu, Wei Lin, Jinrui Yang, Xiawu Zheng, et~al.
\newblock Mme: A comprehensive evaluation benchmark for multimodal large language models.
\newblock \emph{arXiv preprint arXiv:2306.13394}, 2023.

\bibitem[Gao et~al.(2023)Gao, Han, Zhang, Lin, Geng, Zhou, Zhang, Lu, He, Yue, Li, and Qiao]{Gao2023LLaMAAdapterV2}
Peng Gao, Jiaming Han, Renrui Zhang, Ziyi Lin, Shijie Geng, Aojun Zhou, W. Zhang, Pan Lu, Conghui He, Xiangyu Yue, Hongsheng Li, and Yu~Jiao Qiao.
\newblock Llama-adapter v2: Parameter-efficient visual instruction model.
\newblock \emph{ArXiv}, abs/2304.15010, 2023.

\bibitem[Gong et~al.(2023)Gong, Lyu, Zhang, Wang, Zheng, Zhao, Liu, Zhang, Luo, and Chen]{gong2023mmgpt}
Tao Gong, Chengqi Lyu, Shilong Zhang, Yudong Wang, Miao Zheng, Qian Zhao, Kuikun Liu, Wenwei Zhang, Ping Luo, and Kai Chen.
\newblock Multimodal-gpt: A vision and language model for dialogue with humans.
\newblock \emph{arXiv preprint arXiv:2305.04790}, 2023.

\bibitem[Goyal et~al.(2017)Goyal, Khot, Summers{-}Stay, Batra, and Parikh]{balanced_vqa_v2}
Yash Goyal, Tejas Khot, Douglas Summers{-}Stay, Dhruv Batra, and Devi Parikh.
\newblock Making the {V} in {VQA} matter: Elevating the role of image understanding in {V}isual {Q}uestion {A}nswering.
\newblock In \emph{Conference on Computer Vision and Pattern Recognition (CVPR)}, 2017.

\bibitem[He et~al.(2020)He, Fan, Wu, Xie, and Girshick]{he2020momentum}
Kaiming He, Haoqi Fan, Yuxin Wu, Saining Xie, and Ross Girshick.
\newblock Momentum contrast for unsupervised visual representation learning.
\newblock In \emph{Proceedings of the IEEE/CVF conference on computer vision and pattern recognition}, pages 9729--9738, 2020.

\bibitem[Huang et~al.(2023)Huang, Dong, Wang, Hao, Singhal, Ma, Lv, Cui, Mohammed, Liu, Aggarwal, Chi, Bjorck, Chaudhary, Som, Song, and Wei]{Huang2023Kosmos1}
Shaohan Huang, Li Dong, Wenhui Wang, Yaru Hao, Saksham Singhal, Shuming Ma, Tengchao Lv, Lei Cui, Owais~Khan Mohammed, Qiang Liu, Kriti Aggarwal, Zewen Chi, Johan Bjorck, Vishrav Chaudhary, Subhojit Som, Xia Song, and Furu Wei.
\newblock Language is not all you need: Aligning perception with language models.
\newblock \emph{ArXiv}, abs/2302.14045, 2023.

\bibitem[Jiang et~al.(2022)Jiang, Xu, Li, Yan, Ye, Zhang, Bi, and Huang]{jiangtrips}
Chaoya Jiang, Haiyang Xu, Chenliang Li, Ming Yan, Wei Ye, Shikun Zhang, Bin Bi, and Songfang Huang.
\newblock {TRIPS}: Efficient vision-and-language pre-training with text-relevant image patch selection.
\newblock In \emph{Proceedings of the 2022 Conference on Empirical Methods in Natural Language Processing}, pages 4084--4096, Abu Dhabi, United Arab Emirates, 2022. Association for Computational Linguistics.

\bibitem[Jiang et~al.(2023{\natexlab{a}})Jiang, Xu, Ye, Ye, Li, Yan, Bi, Zhang, Huang, and Huang]{jiangbus}
Chaoya Jiang, Haiyang Xu, Wei Ye, Qinghao Ye, Chenliang Li, Ming Yan, Bin Bi, Shikun Zhang, Fei Huang, and Songfang Huang.
\newblock Bus: Efficient and effective vision-language pre-training with bottom-up patch summarization.
\newblock In \emph{Proceedings of the IEEE/CVF International Conference on Computer Vision (ICCV)}, pages 2900--2910, 2023{\natexlab{a}}.

\bibitem[Jiang et~al.(2023{\natexlab{b}})Jiang, Xu, Ye, Ye, Li, Yan, Bi, Zhang, Huang, and Zhang]{jiang2023copa}
Chaoya Jiang, Haiyang Xu, Wei Ye, Qinghao Ye, Chenliang Li, Ming Yan, Bin Bi, Shikun Zhang, Fei Huang, and Ji Zhang.
\newblock Copa: Efficient vision-language pre-training through collaborative object-and patch-text alignment.
\newblock In \emph{Proceedings of the 31st ACM International Conference on Multimedia}, pages 4480--4491, 2023{\natexlab{b}}.

\bibitem[Jiang et~al.(2023{\natexlab{c}})Jiang, Ye, Xu, yan, Zhang, Zhang, and Huang]{Jiang2023VisionLP}
Chaoya Jiang, Wei Ye, Haiyang Xu, Miang yan, Shikun Zhang, Jie Zhang, and Fei Huang.
\newblock Vision language pre-training by contrastive learning with cross-modal similarity regulation.
\newblock In \emph{Annual Meeting of the Association for Computational Linguistics}, 2023{\natexlab{c}}.

\bibitem[Laurençon et~al.(2023)Laurençon, Saulnier, Tronchon, Bekman, Singh, Lozhkov, Wang, Karamcheti, Rush, Kiela, Cord, and Sanh]{laurencon2023idefics}
Hugo Laurençon, Lucile Saulnier, Léo Tronchon, Stas Bekman, Amanpreet Singh, Anton Lozhkov, Thomas Wang, Siddharth Karamcheti, Alexander~M. Rush, Douwe Kiela, Matthieu Cord, and Victor Sanh.
\newblock Obelics: An open web-scale filtered dataset of interleaved image-text documents, 2023.

\bibitem[Li et~al.(2023{\natexlab{a}})Li, Wang, Wang, Ge, Ge, and Shan]{li2023seedbench}
Bohao Li, Rui Wang, Guangzhi Wang, Yuying Ge, Yixiao Ge, and Ying Shan.
\newblock Seed-bench: Benchmarking multimodal llms with generative comprehension.
\newblock \emph{arXiv preprint arXiv:2307.16125}, 2023{\natexlab{a}}.

\bibitem[Li et~al.(2023{\natexlab{b}})Li, Zhang, Chen, Wang, Yang, and Liu]{Li2023Otter}
Bo Li, Yuanhan Zhang, Liangyu Chen, Jinghao Wang, Jingkang Yang, and Ziwei Liu.
\newblock Otter: A multi-modal model with in-context instruction tuning.
\newblock \emph{ArXiv}, abs/2305.03726, 2023{\natexlab{b}}.

\bibitem[Li et~al.(2021)Li, Selvaraju, Gotmare, Joty, Xiong, and Hoi]{li2021align}
Junnan Li, Ramprasaath Selvaraju, Akhilesh Gotmare, Shafiq Joty, Caiming Xiong, and Steven Chu~Hong Hoi.
\newblock Align before fuse: Vision and language representation learning with momentum distillation.
\newblock \emph{Advances in neural information processing systems}, 34:\penalty0 9694--9705, 2021.

\bibitem[Li et~al.(2022)Li, Li, Xiong, and Hoi]{li2022blip}
Junnan Li, Dongxu Li, Caiming Xiong, and Steven Hoi.
\newblock Blip: Bootstrapping language-image pre-training for unified vision-language understanding and generation.
\newblock In \emph{International Conference on Machine Learning}, pages 12888--12900. PMLR, 2022.

\bibitem[Li et~al.(2023{\natexlab{c}})Li, Li, Savarese, and Hoi]{Li2023BLIP2}
Junnan Li, Dongxu Li, Silvio Savarese, and Steven C.~H. Hoi.
\newblock Blip-2: Bootstrapping language-image pre-training with frozen image encoders and large language models.
\newblock \emph{ArXiv}, abs/2301.12597, 2023{\natexlab{c}}.

\bibitem[Li et~al.(2023{\natexlab{d}})Li, Du, Zhou, Wang, Zhao, and rong Wen]{Li2023pope}
Yifan Li, Yifan Du, Kun Zhou, Jinpeng Wang, Wayne~Xin Zhao, and Ji rong Wen.
\newblock Evaluating object hallucination in large vision-language models.
\newblock \emph{ArXiv}, abs/2305.10355, 2023{\natexlab{d}}.

\bibitem[Lin et~al.(2014)Lin, Maire, Belongie, Hays, Perona, Ramanan, Doll{\'a}r, and Zitnick]{lin2014coco}
Tsung-Yi Lin, Michael Maire, Serge Belongie, James Hays, Pietro Perona, Deva Ramanan, Piotr Doll{\'a}r, and C~Lawrence Zitnick.
\newblock Microsoft coco: Common objects in context.
\newblock In \emph{Computer Vision--ECCV 2014: 13th European Conference, Zurich, Switzerland, September 6-12, 2014, Proceedings, Part V 13}, pages 740--755. Springer, 2014.

\bibitem[Liu et~al.(2023{\natexlab{a}})Liu, Lin, Li, Wang, Yacoob, and Wang]{liu2023lrv}
Fuxiao Liu, Kevin Lin, Linjie Li, Jianfeng Wang, Yaser Yacoob, and Lijuan Wang.
\newblock Aligning large multi-modal model with robust instruction tuning.
\newblock \emph{arXiv preprint arXiv:2306.14565}, 2023{\natexlab{a}}.

\bibitem[Liu et~al.(2023{\natexlab{b}})Liu, Lin, Li, Wang, Yacoob, and Wang]{liu2023mitigating}
Fuxiao Liu, Kevin Lin, Linjie Li, Jianfeng Wang, Yaser Yacoob, and Lijuan Wang.
\newblock Mitigating hallucination in large multi-modal models via robust instruction tuning, 2023{\natexlab{b}}.

\bibitem[Liu et~al.(2023{\natexlab{c}})Liu, Li, Li, and Lee]{Liu2023LLava15}
Haotian Liu, Chunyuan Li, Yuheng Li, and Yong~Jae Lee.
\newblock Improved baselines with visual instruction tuning.
\newblock \emph{ArXiv}, abs/2310.03744, 2023{\natexlab{c}}.

\bibitem[Liu et~al.(2023{\natexlab{d}})Liu, Li, Wu, and Lee]{Liu2023Llava}
Haotian Liu, Chunyuan Li, Qingyang Wu, and Yong~Jae Lee.
\newblock Visual instruction tuning.
\newblock \emph{ArXiv}, abs/2304.08485, 2023{\natexlab{d}}.

\bibitem[Liu et~al.(2023{\natexlab{e}})Liu, Duan, Zhang, Li, Zhang, Zhao, Yuan, Wang, He, Liu, et~al.]{liu2023mmbench}
Yuan Liu, Haodong Duan, Yuanhan Zhang, Bo Li, Songyang Zhang, Wangbo Zhao, Yike Yuan, Jiaqi Wang, Conghui He, Ziwei Liu, et~al.
\newblock Mmbench: Is your multi-modal model an all-around player?
\newblock \emph{arXiv preprint arXiv:2307.06281}, 2023{\natexlab{e}}.

\bibitem[Lu et~al.(2022)Lu, Clark, Zellers, Mottaghi, and Kembhavi]{Lu2022UnifiedIO}
Jiasen Lu, Christopher Clark, Rowan Zellers, Roozbeh Mottaghi, and Aniruddha Kembhavi.
\newblock Unified-io: A unified model for vision, language, and multi-modal tasks.
\newblock \emph{ArXiv}, abs/2206.08916, 2022.

\bibitem[Mishra et~al.(2019)Mishra, Shekhar, Singh, and Chakraborty]{mishra2019ocrvqa}
Anand Mishra, Shashank Shekhar, Ajeet~Kumar Singh, and Anirban Chakraborty.
\newblock Ocr-vqa: Visual question answering by reading text in images.
\newblock In \emph{2019 international conference on document analysis and recognition (ICDAR)}, pages 947--952. IEEE, 2019.

\bibitem[Oord et~al.(2018)Oord, Li, and Vinyals]{oord2018representation}
Aaron van~den Oord, Yazhe Li, and Oriol Vinyals.
\newblock Representation learning with contrastive predictive coding.
\newblock \emph{arXiv preprint arXiv:1807.03748}, 2018.

\bibitem[OpenAI(2023{\natexlab{a}})]{2023GPT4VisionSC}
OpenAI.
\newblock Gpt-4v(ision) system card.
\newblock 2023{\natexlab{a}}.

\bibitem[OpenAI(2023{\natexlab{b}})]{OpenAI2023gpt4}
OpenAI.
\newblock Gpt-4 technical report.
\newblock \emph{ArXiv}, abs/2303.08774, 2023{\natexlab{b}}.

\bibitem[Peng et~al.(2023)Peng, Wang, Dong, Hao, Huang, Ma, and Wei]{Peng2023Kosmos2GM}
Zhiliang Peng, Wenhui Wang, Li Dong, Yaru Hao, Shaohan Huang, Shuming Ma, and Furu Wei.
\newblock Kosmos-2: Grounding multimodal large language models to the world.
\newblock \emph{ArXiv}, abs/2306.14824, 2023.

\bibitem[Radford et~al.(2021)Radford, Kim, Hallacy, Ramesh, Goh, Agarwal, Sastry, Askell, Mishkin, Clark, et~al.]{radford2021clip}
Alec Radford, Jong~Wook Kim, Chris Hallacy, Aditya Ramesh, Gabriel Goh, Sandhini Agarwal, Girish Sastry, Amanda Askell, Pamela Mishkin, Jack Clark, et~al.
\newblock Learning transferable visual models from natural language supervision.
\newblock In \emph{International conference on machine learning}, pages 8748--8763. PMLR, 2021.

\bibitem[Schuhmann et~al.(2022)Schuhmann, Beaumont, Vencu, Gordon, Wightman, Cherti, Coombes, Katta, Mullis, Wortsman, et~al.]{schuhmann2022laion}
Christoph Schuhmann, Romain Beaumont, Richard Vencu, Cade Gordon, Ross Wightman, Mehdi Cherti, Theo Coombes, Aarush Katta, Clayton Mullis, Mitchell Wortsman, et~al.
\newblock Laion-5b: An open large-scale dataset for training next generation image-text models.
\newblock \emph{Advances in Neural Information Processing Systems}, 35:\penalty0 25278--25294, 2022.

\bibitem[Singh et~al.(2019)Singh, Natarajan, Shah, Jiang, Chen, Batra, Parikh, and Rohrbach]{singh2019textvqa}
Amanpreet Singh, Vivek Natarajan, Meet Shah, Yu Jiang, Xinlei Chen, Dhruv Batra, Devi Parikh, and Marcus Rohrbach.
\newblock Towards vqa models that can read.
\newblock In \emph{Proceedings of the IEEE/CVF conference on computer vision and pattern recognition}, pages 8317--8326, 2019.

\bibitem[Sun et~al.(2023)Sun, Shen, Cao, Liu, Li, Shen, Gan, Gui, Wang, Yang, Keutzer, and Darrell]{Sun2023LLavaRlhf}
Zhiqing Sun, Sheng Shen, Shengcao Cao, Haotian Liu, Chunyuan Li, Yikang Shen, Chuang Gan, Liangyan Gui, Yu-Xiong Wang, Yiming Yang, Kurt Keutzer, and Trevor Darrell.
\newblock Aligning large multimodal models with factually augmented rlhf.
\newblock \emph{ArXiv}, abs/2309.14525, 2023.

\bibitem[Touvron et~al.(2023{\natexlab{a}})Touvron, Lavril, Izacard, Martinet, Lachaux, Lacroix, Rozi{\`e}re, Goyal, Hambro, Azhar, Rodriguez, Joulin, Grave, and Lample]{Touvron2023LLaMA}
Hugo Touvron, Thibaut Lavril, Gautier Izacard, Xavier Martinet, Marie-Anne Lachaux, Timoth{\'e}e Lacroix, Baptiste Rozi{\`e}re, Naman Goyal, Eric Hambro, Faisal Azhar, Aurelien Rodriguez, Armand Joulin, Edouard Grave, and Guillaume Lample.
\newblock Llama: Open and efficient foundation language models.
\newblock \emph{ArXiv}, abs/2302.13971, 2023{\natexlab{a}}.

\bibitem[Touvron et~al.(2023{\natexlab{b}})Touvron, Martin, Stone, Albert, Almahairi, Babaei, Bashlykov, Batra, Bhargava, Bhosale, Bikel, Blecher, Ferrer, Chen, Cucurull, Esiobu, Fernandes, Fu, Fu, Fuller, Gao, Goswami, Goyal, Hartshorn, Hosseini, Hou, Inan, Kardas, Kerkez, Khabsa, Kloumann, Korenev, Koura, Lachaux, Lavril, Lee, Liskovich, Lu, Mao, Martinet, Mihaylov, Mishra, Molybog, Nie, Poulton, Reizenstein, Rungta, Saladi, Schelten, Silva, Smith, Subramanian, Tan, Tang, Taylor, Williams, Kuan, Xu, Yan, Zarov, Zhang, Fan, Kambadur, Narang, Rodriguez, Stojnic, Edunov, and Scialom]{Touvron2023Llama2}
Hugo Touvron, Louis Martin, Kevin~R. Stone, Peter Albert, Amjad Almahairi, Yasmine Babaei, Nikolay Bashlykov, Soumya Batra, Prajjwal Bhargava, Shruti Bhosale, Daniel~M. Bikel, Lukas Blecher, Cristian~Cant{\'o}n Ferrer, Moya Chen, Guillem Cucurull, David Esiobu, Jude Fernandes, Jeremy Fu, Wenyin Fu, Brian Fuller, Cynthia Gao, Vedanuj Goswami, Naman Goyal, Anthony~S. Hartshorn, Saghar Hosseini, Rui Hou, Hakan Inan, Marcin Kardas, Viktor Kerkez, Madian Khabsa, Isabel~M. Kloumann, A.~V. Korenev, Punit~Singh Koura, Marie-Anne Lachaux, Thibaut Lavril, Jenya Lee, Diana Liskovich, Yinghai Lu, Yuning Mao, Xavier Martinet, Todor Mihaylov, Pushkar Mishra, Igor Molybog, Yixin Nie, Andrew Poulton, Jeremy Reizenstein, Rashi Rungta, Kalyan Saladi, Alan Schelten, Ruan Silva, Eric~Michael Smith, R. Subramanian, Xia Tan, Binh Tang, Ross Taylor, Adina Williams, Jian~Xiang Kuan, Puxin Xu, Zhengxu Yan, Iliyan Zarov, Yuchen Zhang, Angela Fan, Melanie Kambadur, Sharan Narang, Aurelien Rodriguez, Robert Stojnic, Sergey Edunov, and
  Thomas Scialom.
\newblock Llama 2: Open foundation and fine-tuned chat models.
\newblock \emph{ArXiv}, abs/2307.09288, 2023{\natexlab{b}}.

\bibitem[Wang et~al.(2023)Wang, Wu, Han, Peng, Zhong, Zhang, wen Dong, Li, Li, Wang, and He]{Wang2023VIGCVI}
Bin Wang, Fan Wu, Xiao Han, Jiahui Peng, Huaping Zhong, Pan Zhang, Xiao wen Dong, Weijia Li, Wei Li, Jiaqi Wang, and Conghui He.
\newblock Vigc: Visual instruction generation and correction.
\newblock \emph{ArXiv}, abs/2308.12714, 2023.

\bibitem[Yang et~al.(2022)Yang, Duan, Tran, Xu, Chanda, Chen, Zeng, Chilimbi, and Huang]{yang2022triple}
Jinyu Yang, Jiali Duan, Son Tran, Yi Xu, Sampath Chanda, Liqun Chen, Belinda Zeng, Trishul Chilimbi, and Junzhou Huang.
\newblock Vision-language pre-training with triple contrastive learning.
\newblock In \emph{Proceedings of the IEEE/CVF Conference on Computer Vision and Pattern Recognition}, pages 15671--15680, 2022.

\bibitem[Ye et~al.(2023{\natexlab{a}})Ye, Hu, Xu, Ye, Yan, Dan, Zhao, Xu, Li, Tian, Qi, Zhang, and Huang]{mplugdocowl}
Jiabo Ye, Anwen Hu, Haiyang Xu, Qinghao Ye, Ming Yan, Yuhao Dan, Chenlin Zhao, Guohai Xu, Chenliang Li, Junfeng Tian, Qian Qi, Ji Zhang, and Fei Huang.
\newblock mplug-docowl: Modularized multimodal large language model for document understanding.
\newblock \emph{CoRR}, abs/2307.02499, 2023{\natexlab{a}}.

\bibitem[Ye et~al.(2023{\natexlab{b}})Ye, Hu, Xu, Ye, Yan, Xu, Li, Tian, Qian, Zhang, et~al.]{ye2023ureader}
Jiabo Ye, Anwen Hu, Haiyang Xu, Qinghao Ye, Ming Yan, Guohai Xu, Chenliang Li, Junfeng Tian, Qi Qian, Ji Zhang, et~al.
\newblock Ureader: Universal ocr-free visually-situated language understanding with multimodal large language model.
\newblock In \emph{The 2023 Conference on Empirical Methods in Natural Language Processing}, 2023{\natexlab{b}}.

\bibitem[Ye et~al.(2023{\natexlab{c}})Ye, Xu, Xu, Ye, Yan, Zhou, Wang, Hu, Shi, Shi, et~al.]{ye2023mplugowl}
Qinghao Ye, Haiyang Xu, Guohai Xu, Jiabo Ye, Ming Yan, Yiyang Zhou, Junyang Wang, Anwen Hu, Pengcheng Shi, Yaya Shi, et~al.
\newblock mplug-owl: Modularization empowers large language models with multimodality.
\newblock \emph{arXiv preprint arXiv:2304.14178}, 2023{\natexlab{c}}.

\bibitem[Yu et~al.(2023)Yu, Yang, Li, Wang, Lin, Liu, Wang, and Wang]{yu2023mmvet}
Weihao Yu, Zhengyuan Yang, Linjie Li, Jianfeng Wang, Kevin Lin, Zicheng Liu, Xinchao Wang, and Lijuan Wang.
\newblock Mm-vet: Evaluating large multimodal models for integrated capabilities.
\newblock \emph{arXiv preprint arXiv:2308.02490}, 2023.

\bibitem[Zeng et~al.(2021)Zeng, Zhang, and Li]{zeng2021multi}
Yan Zeng, Xinsong Zhang, and Hang Li.
\newblock Multi-grained vision language pre-training: Aligning texts with visual concepts.
\newblock \emph{arXiv preprint arXiv:2111.08276}, 2021.

\bibitem[Zheng et~al.(2023)Zheng, Chiang, Sheng, Zhuang, Wu, Zhuang, Lin, Li, Li, Xing, Zhang, Gonzalez, and Stoica]{zheng2023vicuna}
Lianmin Zheng, Wei-Lin Chiang, Ying Sheng, Siyuan Zhuang, Zhanghao Wu, Yonghao Zhuang, Zi Lin, Zhuohan Li, Dacheng Li, Eric.~P Xing, Hao Zhang, Joseph~E. Gonzalez, and Ion Stoica.
\newblock Judging llm-as-a-judge with mt-bench and chatbot arena, 2023.

\bibitem[Zhu et~al.(2023)Zhu, Chen, Shen, Li, and Elhoseiny]{Zhu2023MiniGPT4}
Deyao Zhu, Jun Chen, Xiaoqian Shen, Xiang Li, and Mohamed Elhoseiny.
\newblock Minigpt-4: Enhancing vision-language understanding with advanced large language models.
\newblock \emph{ArXiv}, abs/2304.10592, 2023.

\end{thebibliography}
